%% file: manuscript-amspreprint.tex
\PassOptionsToPackage{frozencache}{minted}

\IfFileExists{./prepreamble-amspreprint.sty}{\RequirePackage[packages,theorems,changes]{prepreamble-amspreprint}}{}

\makeatletter
\IfFileExists{./scoop-latex/scoop-packages.sty}{\providecommand*{\input@path}{}\edef\input@path{{./scoop-latex/}\input@path}}{}
\makeatother
\PassOptionsToPackage{sorting = none, style = nature, backend = biber}{biblatex}
\PassOptionsToPackage{commentmarkup = footnote}{changes}
\PassOptionsToPackage{frozencache}{minted}    

\documentclass[english]{amsart}

\RequirePackage{biblatex}
\RequirePackage{ltxcmds}
\IfFileExists{./preamble-amspreprint.sty}{\RequirePackage[packages,theorems,changes]{preamble-amspreprint}}{}

\usepackage{manuscript}

\IfFileExists{./postpreamble-amspreprint.sty}{\RequirePackage[packages,theorems,changes]{postpreamble-amspreprint}}{}

\makeatletter
\@ifpackageloaded{changes}{
\definechangesauthor[name = {Viktor Martinek}, color = {red!80!black}]{VM}
}{}
\makeatother        

\makeatletter
\@ifpackageloaded{hyperref}{%
	\hypersetup{
		pdftitle = {Fast Symbolic Regression Benchmarking},
		pdfauthor = {Viktor Martinek},
		pdfkeywords = {symbolic regression, benchmark, hyperparameter tuning},
		pdfcreator = {Created using the Scoop Template Engine version 1.6.0.}
	}
}{
	\pdfinfo{
		/Title (Fast Symbolic Regression Benchmarking)
		/Author (Viktor Martinek)
		/Subject ()
		/Keywords (symbolic regression, benchmark, hyperparameter tuning)
		/Creator (Created using the Scoop Template Engine version 1.6.0.)
	}
}
\makeatother

\title[Fast Symbolic Regression Benchmarking]{Fast Symbolic Regression Benchmarking}

\author[V. Martinek]{Viktor Martinek\orcidlink{0000-0001-6215-4783}}
\address[V. Martinek]{Interdisciplinary Center for Scientific Computing, Heidelberg University, 69120 Heidelberg, Germany}
\email{\detokenize{viktor.martinek@iwr.uni-heidelberg.de}}

\thanks{This work was funded by the Deutsche Forschungsgemeinschaft (DFG, German Research Foundation) -- HE~6077/14-2 -- within the Priority Programme \enquote{SPP 2331: Machine Learning in Chemical Engineering}.}

\date{\today}

\dedicatory{}

\begin{document}

\begin{abstract}
\input{abstract.tex}
\end{abstract}


\makeatletter
\ltx@ifpackageloaded{hyperref}{%
\subjclass[2010]{}
}{%
\subjclass[2010]{}
}
\makeatother

\maketitle

\input{content.tex}

\appendix
\input{appendix.tex}

\printbibliography

\end{document}

%% file: abstract.tex

\Ac{SR} uncovers mathematical models from data.
Several benchmarks have been proposed to compare the performance of \ac{SR} algorithms.
However, existing ground-truth rediscovery benchmarks overemphasize the recovery of \enquote{the one} expression form or rely solely on computer algebra systems (such as \sympy) to assess success.
Furthermore, existing benchmarks continue the expression search even after its discovery.
We improve upon these issues by introducing curated lists of acceptable expressions, and a callback mechanism for early termination.
As a starting point, we use the \ac{SRSD} benchmark problems proposed by Yoshitomo~\etal, and benchmark the two \ac{SR} packages \symbolicregressionjl and \tisr.
The new benchmarking method increases the rediscovery rate of \symbolicregressionjl from \SI[round-mode = places, round-precision = 1]{26.7}{\percent}, as reported by Yoshitomo~\etal, to \SI[round-mode = places, round-precision = 1]{44.7}{\percent}.
Performing the benchmark takes \SI[round-mode = places, round-precision = 1]{41.2}{\percent} less computational expense.
\tisr's rediscovery rate is \SI[round-mode = places, round-precision = 1]{69.4}{\percent}, while performing the benchmark saves \SI[round-mode = places, round-precision = 0]{63}{\percent} time.

%% file: content.tex

\acresetall

\section{Introduction}
\label{section:introduction}

\Ac{SR} aims to discover mathematical expressions that fit data without assuming a predefined model, making it a powerful tool for scientific discovery.
In contrast to many other supervised machine learning approaches, \ac{SR} aims to discover the expression structure and identify its parameters.
There is much research and many types of algorithms for \ac{SR}.
However, as proven by \cite{VirgolinPissis:2022:1}, \ac{SR} is NP-hard and very computationally expensive.

Its optimization objectives typically balance two goals: achieving a high fit quality to accurately model the data, and minimizing expression complexity to ensure interpretability.
The complexity reflects an expression’s structural simplicity.
It is commonly measured by the number of operators and operands, favoring concise expressions.

Realistic benchmarking is crucial, as it enables fair and meaningful comparisons between methods by testing them under conditions that mirror practical applications.
It also facilitates hyperparameter tuning by providing a representative set of challenges, ensuring optimized settings that translate to real-world performance.
\ac{SR} benchmarking primarily falls into two categories: ground-truth rediscovery and real-world.
Ground-truth rediscovery benchmarks use datasets with known expressions to test an algorithm’s ability to rediscover those.
Some benchmarks also utilize a fallback measure, \eg, the coefficient of determination, to judge the fit quality, or the normalized editing distance, to judge the similarity to the ground truth expression.
Real-world benchmarks, conversely, employ experimental or observational data without known expressions, evaluating the capacity to uncover novel, interpretable expressions.

Requiring exact rediscovery of an expression does neither fully align with \ac{SR}'s optimization objectives, nor does it reflect a real-world use case.
Equivalent forms of the expression should be considered found if they are functionally equivalent and remain concise.
For many expressions, there are potentially infinite functionally equivalent forms with the same or a similar complexity.
As a first example, a list of expressions is provided below.
All listed expressions are functionally equivalent forms of the \ac{SRSD} benchmark problem \texttt{II.38.14}.

\begin{itemize}
    \item problem ID \texttt{II.38.14}
        \begin{multicols}{3}
            \begin{itemize}
                \item $v_1 / (2 \cdot (1 + v_2))$
                \item $v_1 / (2 + 2 v_2)$
                \item $0.5 v_1 / (v_2 + 1)$
            \end{itemize}
        \end{multicols}
\end{itemize}

In the above example, the functionally equivalent forms consist of the same number of operators and operands.
Thus, there is no difference in their complexity, and no incentive for the \ac{SR} algorithm to prefer any of these forms.

La~Cava~\etal \cite{LaCavaOrzechowskiBurlacuOlivettiDeFrancaVirgolinJinKommendaMoore:2021:1} introduced a new definition for a symbolic solution, which aims to detect functionally equivalent expressions.
Both, the target and the candidate expression, are simplified using a computer algebra system, \ie the \python (see \cite{van1995python}) package \sympy (see \cite{sympy}).
If, according to \sympy, their difference or their ratio simplifies to a constant, the expressions are deemed equivalent.
This definition accepts any constant, not only zero or one for the difference and the ratio, respectively.
The ability of this approach to detect expressions, which differ at most by a constant offset or multiplier, is limited by the capabilities of the \sympy library.
For the first example shown above, this approach works.
However, this approach fails in some cases, such as the following examples from the \ac{SRSD} benchmark.
The constants in the expressions listed below are rounded to five significant digits.

\begin{itemize}
    \item problem ID \texttt{I.34.1}
        \begin{multicols}{2}
            \begin{itemize}
                \item $-v_1/(3.3356 \cdot 10^{-9} \cdot v_2-1)$
                \item $-2.9979 \cdot 10^{8} v_1/(v_2-2.9979 \cdot 10^{8})$
            \end{itemize}
        \end{multicols}
    \item problem ID \texttt{II.11.3}
        \begin{multicols}{2}
            \begin{itemize}
                \item $v_1 v_2/(v_3 \cdot (v_4^2-v_5^2))$
                \item $v_1 v_2/(v_3 \cdot (v_4-v_5) \cdot (v_4+v_5))$
            \end{itemize}
        \end{multicols}
    \item problem ID \texttt{I.6.2a}
        \begin{multicols}{2}
            \begin{itemize}
                \item $0.39894 \cdot \exp(-v_1^2 / 2)$
                \item $0.39894 \cdot \sqrt{\exp(-v_1^2)}$
                \item $0.39894 / \sqrt{\exp(v_1^2)}$
                \item $0.39894 \cdot 0.60653^{v_1^2}$
            \end{itemize}
        \end{multicols}
\end{itemize}

All expression forms shown above should be equally accepted.
In a real-world setting, a researcher utilizing \ac{SR} and finding an optimally-matching, concise expression, can and would rearrange it to their preferences or use-case.

On the other hand, \ac{SR} is very computationally expensive.
\ac{SR} benchmarks allot a high computational budget for algorithms.
This budget is fixed for each run, and the search continues even after the expression is discovered.
As \ac{SR} algorithms are typically non-deterministic, each problem is run several times, multiplying the impact of this inefficiency.
Also, many \ac{SR} algorithms, in particular, genetic programming-based approaches, have many hyperparameters, which must be tuned for optimal performance.
Thus, tuning the hyperparameters of \ac{SR} algorithms is very expensive.

In the next section, we provide a brief overview of some existing \ac{SR} benchmarks.
In \cref{sec:method}, we introduce our proposed method, and in \cref{sec:experiments}, we demonstrate its effectiveness.
Lastly, we draw a conclusion in \cref{sec:conclusion}.

\section{Related Work}
\label{sec:relatedwork}

The Feynman \ac{SR} benchmark (see \cite{UdrescuTegmark:2020:1}), derived from the Feynman Lectures on Physics, provides a set of expressions to test an algorithm's ability to recover known physical laws.
For all variables and constants occurring in the expressions, sampling ranges are provided, which are, however, not realistic.

\srbench (see \cite{LaCavaOrzechowskiBurlacuOlivettiDeFrancaVirgolinJinKommendaMoore:2021:1}) offers a broader evaluation framework, encompassing both synthetic and real-world datasets to assess \ac{SR} methods across diverse conditions.
Among other things, they adopt the Feynman benchmark problems, and extend them with real-world datasets without known ground truth.
As mentioned in the previous section, in their ground-truth rediscovery benchmarks, they relax the requirement for an exact rediscovery using a new definition of a symbolic solution.
If, according to \sympy, the difference or the ratio of two expressions simplify to a constant, the expressions are deemed equivalent.

\empiricalbench (see \cite{Cranmer:2023:1}) focuses on real-world applications.
Although the ground-truth expressions are known, they are to be discovered based on realistic datasets, \ie, the variable ranges are realistic.
And, even more importantly, the constants are not provided as variables, but need to be identified by the \ac{SR} algorithm.
To assess whether the discovered expressions are equivalent to the ground-truth, they manually inspect the hall of fame\footnote{set of best (non-dominated) expressions found during the expression search} at the end of the search.
This approach is thorough, but time-consuming and error-prone.

The \ac{SRSD} (see \cite{matsubara2024rethinking}) benchmark adopts the expressions from the Feynman \ac{SR} benchmark, and defines realistic sampling ranges for the variables.
For the synthetic data generation, the real values of the physics constants are used, and have to be identified by the \ac{SR} algorithm.
To assess rediscovery, they adopt \srbench's approach.

\section{Method}
\label{sec:method}

In this work, two challenges of ground-truth rediscovery benchmarks are addressed, namely, the acceptance of functionally equivalent expressions, and the high temporal expense of conducting such benchmarks.
We propose to define a curated list of acceptable forms for each expression, and a callback mechanism for early termination.

The callback mechanism is a function, which is invoked periodically during the \ac{SR} search, \ie, in approximately \SI[round-mode = figures, round-precision = 2]{15}{\second} intervals.
Promising candidate expressions, \eg, the expressions in the hall of fame, are processed and compared to the acceptable expressions list.
As processing a candidate expression incurs a computational expense, it must satisfy a set of criteria before processing.
We require the maximum relative error on test data to be below a set threshold, \ie, \SI[round-mode = figures, round-precision = 1]{0.000001}{\percent}.
Next, the expression is simplified using \sympy, and its constants are rounded to a set number of significant digits, \ie five.
If a processed expression matches one in the list of acceptable expressions, the search is terminated with success.

If the processed expression does not match any in the list, it is recorded separately without terminating the search.
During post-processing, these recorded expressions are inspected for new functionally equivalent expressions.
Those are retroactively accepted, and added to the list of acceptable expressions for future runs.
This step ensures that manual inspection of expressions, as conducted in \empiricalbench, is reused.
Thus, the efficiency of this benchmark increases with use.

It would also be possible to use the aforementioned relaxed definition of a symbolic solution to compare candidate expressions with the list of acceptable expressions.
However, our trials show that this approach is prohibitively expensive to be used for the proposed live monitoring. 
The overhead introduced by the simplification and parameter rounding only, on the other hand, is acceptable.

Even if the list of acceptable expressions changes, fairness is ensured, as long as the rules of acceptable expressions remain the same.
The reference expression is the original expression simplified by \sympy.
On top of being algebraically equivalent, we propose that number of operators and operands should be at most \SI[round-mode = figures, round-precision = 1]{20}{\percent} higher than the reference expression (using binary operators). 
The pseudo-algorithm for the callback is shown in \cref{alg:1}.

\begin{algorithm}
    \caption{Callback for Early Termination}
    \begin{algorithmic}[1]
        \Require candidate expression $e$,
                 fit quality of $e$ on test data $\delta(e)$,
                 number of operators and operands in $e$ $c(e)$,
                 fit quality threshold on test data $\delta_{\mathrm{max}}$,
                 threshold number of operators and operands $c_{\mathrm{max}}$,
                 acceptable expressions $A$ with parameters rounded to $n$ significant digits

        \Function{callback}{$c$}
            \If {$\delta(e) \geq \delta_{\mathrm{max}}$}
                \State \Return false
            \EndIf

            \State $e \gets \operatorname{simplify}(e)$ \Comment {using \sympy}
            \State $e \gets \operatorname{round\_constants}(e)$ \Comment {round constants to $n$ significant digits}

            \If {$e \in A$}
                \State record discovery, e.g., in a text file
                \State \Return true
            \EndIf

            \State \Comment {filtering unpromising candidates, e.g., required variables must occur in $c$}
            \If {additional prerequisites are met}
                \State record potential acceptable expression form, e.g., in a text file
            \EndIf

            \State \Return false
        \EndFunction
    \end{algorithmic}
    \label{alg:1}
\end{algorithm}

We highlight that the validity of early termination depends on the assumption that the expression triggering the termination, or a different acceptable expression, would be present in the hall of fame at the end of the allotted time or computational budget.
The expression would be displaced from the hall of fame is if a superior solution, as defined by the hall of fame’s selection criteria, replaces it.
In the absence of noise, and if the hall of fame’s selection criteria are perfectly aligned with a solution's utility, any such superior solution should also be deemed acceptable. 

\section{Experiments}
\label{sec:experiments}

In this section, we discuss the experimental setup and the results of the proposed benchmarking method with the two \ac{SR} packages \symbolicregressionjl (see \cite{Cranmer:2023:1}) and \tisr (see \cite{MartinekFrotscherRichterHerzog:2023:1,martinek202514800761}).
We choose \symbolicregressionjl to compare its performance on this benchmark to the one conducted by \cite{matsubara2024rethinking}, and \tisr as a second \julia \ac{SR} tool.

\subsection{Setup}
\label{sec:setup}

We use the $120$~benchmark problems defined in the \ac{SRSD} benchmark (see \cite{matsubara2024rethinking}).
No fallback method is employed, \ie, only rediscovery of the expressions, as defined in the previous section, is reported.
Each expression is searched five times allowing a maximum search time of \SI[round-mode = figures, round-precision = 1]{30}{\minute} for each run.
The data are sampled according to the specifications of \cite{matsubara2024rethinking}.
For each run, we sample $200$ points for the training and the test dataset, respectively.
Both algorithms are configured to not use any multiprocessing.
We conduct the experiments on a machine equipped with an AMD Ryzen 7 PRO 5850U with Radeon Graphics, which has eight physical cores, and $12$~GiB of RAM.
The \texttt{GNU parallel} library (see \cite{gnuparallel}) is utilized to run at most eight jobs in parallel.
Each benchmark problem is run five times within one job, and a timeout of \SI[round-mode = figures, round-precision = 1]{10000}{\second} is set for each job.

For each of the benchmark problems, we adapt the maximum allowed expression complexity to \SI[round-mode = figures, round-precision = 1]{50}{\percent} higher than the reference.
This is in contrast allowing a fixed maximal complexity for all benchmark problems, and how \cite{matsubara2024rethinking} conducted the experiments.
This greatly increases efficiency at the cost of introducing some bias.

The function set\footnote{available operators to construct the expression from} for all benchmark problems consists of: \texttt{+}, \texttt{-}, \texttt{*}, \texttt{/}, \texttt{pow}, \texttt{neg}, \texttt{exp}, \texttt{log}, \texttt{sqrt}, \texttt{pow2}, \texttt{pow3}, \texttt{sin}, \texttt{cos}, \texttt{tanh}. 
Some operator nesting is prohibited.
For both \ac{SR} libraries, \texttt{log} and \texttt{exp} cannot be nested with themselves or each other.
Likewise for \texttt{sin}, \texttt{cos}, and \texttt{tanh}.
In addition, in \tisr, the power operator \texttt{pow} can only raise to the power a constant.

The callback functions are implemented along the lines of \cref{alg:1}, except for some additional data structure transformations and canonicalizations.
To reliably determine the relative error $\delta_{max}$, it is calculated using an offset $o = 10^{-100}$:
\begin{equation*}
    \delta_{max}
    =
    \frac{\|y_{\mathrm{orig}} - y_{\mathrm{pred}}\|}{\|y_{\mathrm{orig}}\| + o} ,
\end{equation*}
where $y_{\mathrm{orig}}$ and $y_{\mathrm{pred}}$ are the target and the prediction, respectively.
The callback function is called in approximately \SI[round-mode = figures, round-precision = 2]{15}{\second} intervals for all expressions in the hall of fame.

\tisr supports the required callback functionality with the utilized commit \texttt{5b541b3}.
To implement the callback in \symbolicregressionjl, it is cloned at commit \texttt{3a85b696}, and the existing callback functionality adapted to accommodate this benchmark's requirements.
The \texttt{git diff} between the cloned state and the one utilized for this work is shown in \cref{sec:symregdiff}.
The code to repeat the experiments, as well as the processed output data generated for this publication, can be found at \url{https://github.com/viktmar/FastSRB-paper-repo}.

\subsection{Results}
\label{sec:results}

The rediscovery rates of \symbolicregressionjl in the benchmark conducted by \cite{matsubara2024rethinking}, as well as its, and \tisr's rediscovery rate using the proposed benchmarking method, are shown in \cref{table:rediscovery}.
Benchmark problems \texttt{B4}, \texttt{B11}, and \texttt{III.9.52} are omitted from all statistics, due to repeated \sympy issues during the experiments.

\begin{table}[htb]
	\centering
    \caption{The rediscovery rates of \symbolicregressionjl (\texttt{PySR} in the table) as reported by \cite{matsubara2024rethinking}, as well as its, and \tisr's rediscovery rate using the proposed benchmarking method (\fastsrb). The results are binned according to the \enquote{easy}, \enquote{medium}, and \enquote{hard} categories proposed by \cite{matsubara2024rethinking}. The results are averaged within the categories and five runs for each problem. Benchmark problems \texttt{B4}, \texttt{B11}, and \texttt{III.9.52} are omitted from all statistics, due to repeated \sympy issues during the experiments.}
    \begin{tabular}{@{}p{0.15\textwidth}@{}p{0.25\textwidth}@{}p{0.25\textwidth}@{}p{0.25\textwidth}}
		\toprule
        \textbf{Category} & \texttt{PySR} before                      & \texttt{PySR} with \fastsrb                                                 & \tisr with \fastsrb                                             \\
		\midrule
        Easy              & \SI[round-mode = places, round-precision = 1]{60.0}{\percent} & \SI[round-mode = places, round-precision = 1]{86.0}{\percent} & \SI[round-mode = places, round-precision = 1]{100.0}{\percent} \\
        Medium            & \SI[round-mode = places, round-precision = 1]{30.0}{\percent} & \SI[round-mode = places, round-precision = 1]{46.0}{\percent} & \SI[round-mode = places, round-precision = 1]{78.0}{\percent}  \\
        Hard              & \SI[round-mode = places, round-precision = 1]{4.0}{\percent}  & \SI[round-mode = places, round-precision = 1]{17.4}{\percent} & \SI[round-mode = places, round-precision = 1]{42.6}{\percent}  \\
		\bottomrule
	\end{tabular}
	\label{table:rediscovery}
\end{table}

\begin{figure}[hbt]
    \centering
	\includegraphics[width=0.9\textwidth]{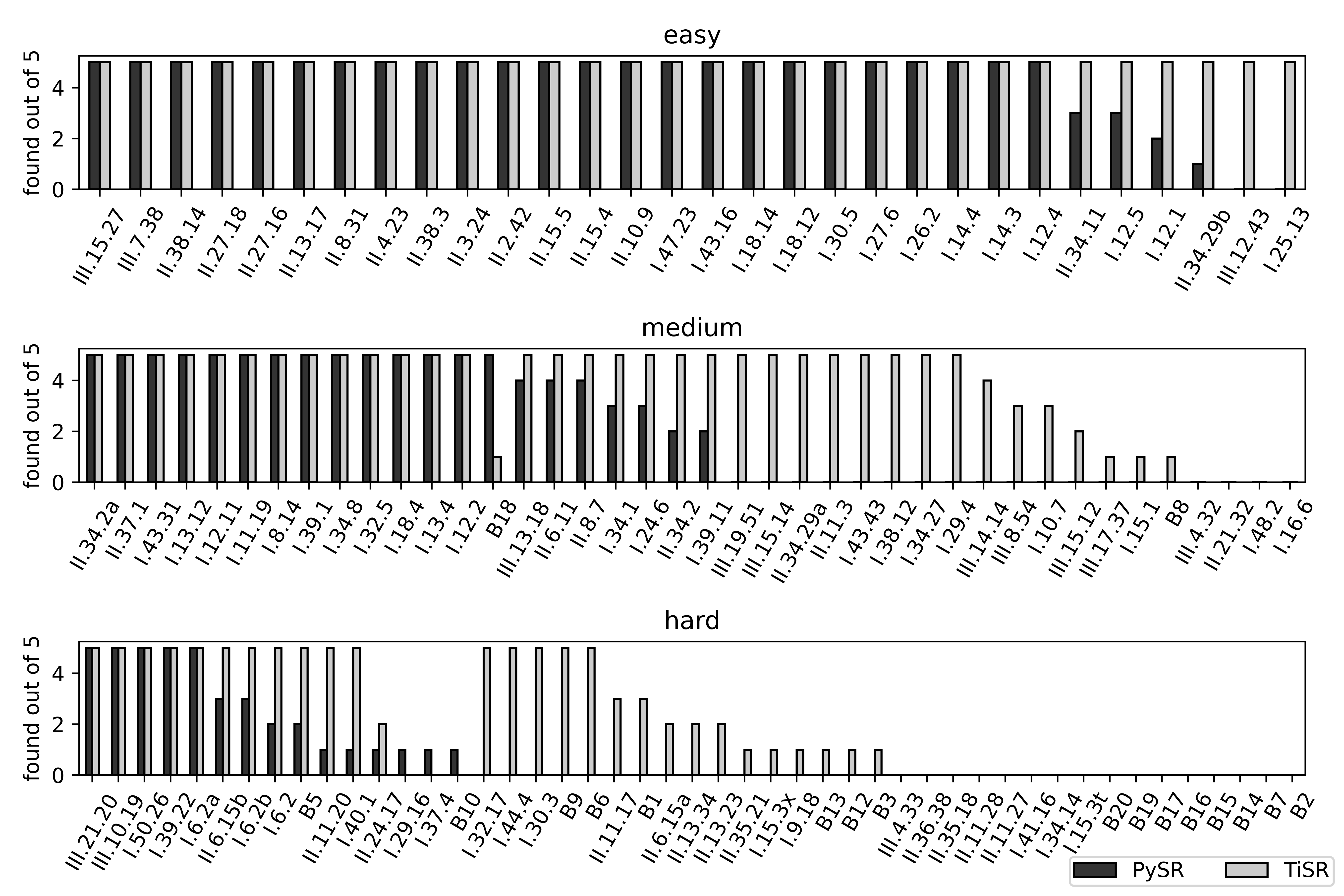}
    \caption{The rediscovery rates of \symbolicregressionjl (\texttt{PySR} in the figure legend) and \tisr using the proposed benchmarking method (\fastsrb) for the benchmark problems defined by \cite{matsubara2024rethinking}. The results are shown in three figures, gathering problems according to the \enquote{easy}, \enquote{medium}, and \enquote{hard} categories proposed by \cite{matsubara2024rethinking}. Benchmark problems \texttt{B4}, \texttt{B11}, and \texttt{III.9.52} are omitted from all statistics, due to repeated \sympy issues during the experiments.}
	\label{fig:foundplot}
\end{figure}

As can be seen in \cref{table:rediscovery}, the rediscovery rates of \symbolicregressionjl are significantly higher using the proposed method.
The overall rediscovery rate increases from \SI[round-mode = places, round-precision = 1]{26.7}{\percent} to \SI[round-mode = places, round-precision = 1]{44.7}{\percent}.
\tisr's overall rediscovery rate is \SI[round-mode = places, round-precision = 1]{69.4}{\percent}.
The rediscovery results of \symbolicregressionjl and \tisr for each benchmark problem using the \fastsrb method are shown in \cref{fig:foundplot}.

In terms of time, the early termination feature saved \SI[round-mode = places, round-precision = 1]{41.2}{\percent} of computational expense for the complete run using \symbolicregressionjl, which amounts to a total of \SI[round-mode = places, round-precision = 0]{120}{\hour}.
\tisr saved \SI[round-mode = places, round-precision = 0]{63}{\percent} of computational expense, amounting to \SI[round-mode = places, round-precision = 0]{184}{\hour}.

All acceptable expressions at the time of publication for an excerpt of the problems are shown in \cref{sec:acceptables}.

\section{Conclusion}
\label{sec:conclusion}

In this work, a new method for ground-truth rediscovery \ac{SR} benchmarks is proposed, which reflects realistic use cases for \ac{SR} more closely, and increases benchmarking efficiency.
The requirements for rediscovery are relaxed to allow functionally equivalent forms of an expression meeting several criteria.
Curated lists of acceptable expression forms are introduced, which are continually extended if new forms are discovered.
The benchmarking approach is implemented and tested using the \ac{SRSD} benchmark problems, and the two packages \symbolicregressionjl and \tisr are benchmarked.
This new approach increases the overall rediscovery rate of \symbolicregressionjl from \SI[round-mode = places, round-precision = 1]{26.7}{\percent}, as reported by \cite{matsubara2024rethinking}, to \SI[round-mode = places, round-precision = 1]{44.7}{\percent} on the same set of benchmark problems.
Furthermore, a live-monitoring approach is proposed, which terminates expression search early, if a known acceptable form is discovered.
For \symbolicregressionjl, the early termination saved \SI[round-mode = places, round-precision = 1]{41.2}{\percent} of computational expense for the complete benchmark run.
\tisr's overall rediscovery rate is \SI[round-mode = places, round-precision = 1]{69.4}{\percent}, while \SI[round-mode = places, round-precision = 0]{63}{\percent} of computational expense are saved using the early termination.
A \julia package containing the benchmark problems, the acceptable expression forms, sampling functionality, as well as convenience functions for processing the expressions and designing the callback function, is provided at \url{https://github.com/viktmar/FastSRB}.

%% file: appendix.tex

\section{Changes to \symbolicregressionjl}
\label{sec:symregdiff}

In the following, the \texttt{git diff} between the cloned state at commit \texttt{3a85b696}, and the utilized state is shown.

\begin{minted}
[
frame=lines,
framesep=2mm,
baselinestretch=1.2,
bgcolor=light-gray,
fontsize=\footnotesize,
linenos,
breaklines
]
{diff}
M src/SearchUtils.jl
@@ -354,10 +354,16 @@ end
 function _check_for_loss_threshold(_, ::Nothing, ::AbstractOptions)
     return false
 end
+
+global last_called = time()
+
 function _check_for_loss_threshold(halls_of_fame, f::F, options::AbstractOptions) where {F}
+    time() - last_called > 15 || return false
+    global last_called = time()
+
     return all(halls_of_fame) do hof
         any(hof.members[hof.exists]) do member
-            f(member.loss, compute_complexity(member, options))::Bool
+            f(member, options)::Bool
         end
     end
 end
\end{minted}

\section{Acceptable Expressions}
\label{sec:acceptables}

In the following, all acceptable expression forms for an excerpt of the benchmark problems is provided.
The parameters are rounded to five significant digits.

\begin{itemize}
    \item problem ID \texttt{I.47.23}
        \begin{itemize}
            \begin{multicols}{2}
                \item \texttt{sqrt(((v1*v2)/v3))}
                \item \texttt{(sqrt(v2)*sqrt((v1/v3)))}
                \end{multicols}
        \end{itemize}
    \item problem ID \texttt{II.3.24}
        \begin{itemize}
            \begin{multicols}{2}
                \item \texttt{((0.079577*v1)/(v2\string^2))}
                \item \texttt{(0.079577*v1*sqrt((v2\string^-4)))}
                \end{multicols}
        \end{itemize}
    \item problem ID \texttt{I.18.4}
        \begin{itemize}
            \begin{multicols}{2}
                \item \texttt{(((v1*v2)+(v3*v4))/(v1+v3))}
                \item \texttt{(v2-((v2-v4)/((v1/v3)+1)))}
                \item \texttt{((1*((v1*v2)+(v3*v4)))/(v1+v3))}
                \end{multicols}
        \end{itemize}
    \item problem ID \texttt{I.24.6}
        \begin{itemize}
            \item \texttt{(0.25*v1*(v4\string^2)*((v2\string^2)+(v3\string^2)))}
            \item \texttt{(0.25*v1*(v4\string^2)*((1*(v2\string^2))+(v3\string^2)))}
            \item \texttt{(v1*(v4\string^2)*((0.25*(v2\string^2))+(0.25*(v3\string^2))))}
            \item \texttt{(0.25*v1*(v4\string^2)*((v2\string^2)+(1*(v3\string^2))))}
            \item \texttt{(0.25*(v1\string^1)*(v4\string^2)*((v2\string^2)+(v3\string^2)))}
            \item \texttt{(v1*(v4\string^2)*((0.25*(v2\string^2))+(0.25*(v3\string^2))))}
        \end{itemize}
    \item problem ID \texttt{I.43.43}
        \begin{itemize}
            \item \texttt{((1.3806e-23*v2)/(v3*(v1-1)))}
            \item \texttt{(v2/(v3*((7.243e22*v1)-7.243e22)))}
        \end{itemize}
    \item problem ID \texttt{II.21.32}
        \begin{itemize}
            \item \texttt{((-8.9877e9*v1)/(v2*((3.3356e-9*v3)-1)))}
            \item \texttt{((-2.6945e18*v1)/(v2*(v3-2.9979e8)))}
        \end{itemize}
    \item problem ID \texttt{I.8.14}
        \begin{itemize}
            \item \texttt{sqrt((((v1-v2)\string^2)+((v3-v4)\string^2)))}
            \item \texttt{((((v1-v2)\string^2)+((v3-v4)\string^2))\string^0.5)}
        \end{itemize}
    \item problem ID \texttt{I.29.16}
        \begin{itemize}
            \item \texttt{sqrt((((v1\string^2)-(2*v1*v2*cos((v3-v4))))+(v2\string^2)))}
            \item \texttt{sqrt(((-(v1)*v2*((2*cos((v3-v4)))+2))+((v1+v2)\string^2)))}
            \item \texttt{sqrt(((-(v1)*v2*((2*cos((v3-v4)))-2))+((v1-v2)\string^2)))}
            \item \texttt{sqrt(((v1*(v1-(2*v2*cos((v3-v4)))))+(v2\string^2)))}
            \item \texttt{sqrt(((-2*v1*v2*cos((v3-v4)))+(v1\string^2)+(v2\string^2)))}
            \item \texttt{(1.4142*sqrt(((-(v1)*v2*(cos((v3-v4))-1))+(0.5*((v1-v2)\string^2)))))}
            \item \texttt{sqrt(((v1\string^2)-(v2*((2*v1*cos((v3-v4)))-v2))))}
            \item \texttt{sqrt(((v1\string^2)-(v2*((2*v1*sin((-(v3)+v4+1.5708)))-v2))))}
        \end{itemize}
    \item problem ID \texttt{I.37.4}
        \begin{itemize}
            \item \texttt{(v1+v2+(2*sqrt((v1*v2))*cos(v3)))}
            \item \texttt{((2*sqrt(v1)*sqrt(v2)*cos(v3))+v1+v2)}
        \end{itemize}
    \item problem ID \texttt{I.44.4}
        \begin{itemize}
            \item \texttt{(1.3806e-23*v1*v2*log((v3/v4)))}
            \item \texttt{(-1.3806e-23*v1*v2*log((v4/v3)))}
            \item \texttt{(1.3806e-23*v1*v2*(log(v3)-log(v4)))}
            \item \texttt{(-1.3806e-23*v1*v2*(log(v4)-log(v3)))}
        \end{itemize}
    \item problem ID \texttt{II.24.17}
        \begin{itemize}
            \item \texttt{(3.1416*sqrt(((1.1274e-18*(v1\string^2))-(1/(v2\string^2)))))}
            \item \texttt{((3.1416*sqrt(((1.1274e-18*(v1\string^2)*(v2\string^2))-1)))/v2)}
        \end{itemize}
    \item problem ID \texttt{B1}
        \begin{itemize}
            \item \texttt{((3.3266000000000004e-57*(v1\string^2)*(v2\string^2))/((v3\string^2)*(sin((v4/2))\string^4)))}
            \item \texttt{((3.3266000000000004e-57*(v1\string^2)*(v2\string^2))/((v3\string^2)*(sin((0.5*v4))\string^4)))}
            \item \texttt{((1.3307e-56*(v1\string^2)*(v2\string^2))/((v3\string^2)*((cos(v4)-1)\string^2)))}
        \end{itemize}
    \item problem ID \texttt{B6}
        \begin{itemize}
            \item \texttt{sqrt((((2*(v1\string^2)*v2*(v3\string^2))/(v4*(v5\string^2)*(v6\string^2)*(v7\string^4)))+1))}
            \item \texttt{(1.4142*sqrt(((((v1\string^2)*v2*(v3\string^2))/(v4*(v5\string^2)*(v6\string^2)*(v7\string^4)))+0.5)))}
        \end{itemize}
    \item problem ID \texttt{B13}
        \begin{itemize}
            \item \texttt{((0.079577*v2)/(v1*sqrt((((v3\string^2)-(2*v3*v4*cos(v5)))+(v4\string^2)))))}
            \item \texttt{((0.079577*v2*((1/(((v3\string^2)-(2*v3*v4*cos(v5)))+(v4\string^2)))\string^0.5))/v1)}
        \end{itemize}
    \item problem ID \texttt{B20}
        \begin{itemize}
            \item \texttt{((2.4946e-29*v1*(((v1\string^2)-(v1*v2*(sin(v3)\string^2)))+(v2\string^2)))/(v2\string^3))}
            \item \texttt{((2.4946e-29*v1*((v1*(v1-(v2*(sin(v3)\string^2))))+(v2\string^2)))/(v2\string^3))}
        \end{itemize}
\end{itemize}